\documentclass[conference]{IEEEtran}
\IEEEoverridecommandlockouts
\usepackage{cite}
\usepackage{amsmath,amssymb,amsfonts}
\usepackage{algorithmic}
\usepackage{graphicx}
\usepackage{textcomp}
\usepackage{xcolor}

\usepackage{subcaption}
\usepackage{tabularx}
\usepackage{booktabs}
\usepackage{dcolumn}
\usepackage{hyperref}

\def\BibTeX{{\rm B\kern-.05em{\sc i\kern-.025em b}\kern-.08em
    T\kern-.1667em\lower.7ex\hbox{E}\kern-.125emX}}
\begin{document}

\title{Towards Efficient Cross-Modal Visual Textual Retrieval using Transformer-Encoder Deep Features}

\author{\IEEEauthorblockN{Nicola Messina, Giuseppe Amato, Fabrizio Falchi, Claudio Gennaro}
\IEEEauthorblockA{\textit{Institute of Information Science and Technologies, National Research Council} \\
Pisa, Italy \\
\{name.surname\}@isti.cnr.it}
\and
\IEEEauthorblockN{Stéphane Marchand-Maillet}
\IEEEauthorblockA{\textit{University of Geneva} \\
Geneva, Switzerland \\
stephane.marchand-maillet@unige.ch}
}

\maketitle

\begin{abstract}
Cross-modal retrieval is an important functionality in modern search engines, as it increases the user experience by allowing queries and retrieved objects to pertain to different modalities.
In this paper, we focus on the image-sentence retrieval task, where the objective is to efficiently find relevant images for a given sentence (image-retrieval) or the relevant sentences for a given image (sentence-retrieval). 
Computer vision literature reports the best results on the image-sentence matching task using deep neural networks equipped with attention and self-attention mechanisms. They evaluate the matching performance on the retrieval task by performing sequential scans of the whole dataset. This method does not scale well with an increasing amount of images or captions.
In this work, we explore different preprocessing techniques to produce sparsified deep multi-modal features extracting them from state-of-the-art deep-learning architectures for image-text matching. Our main objective is to lay down the paths for efficient indexing of complex multi-modal descriptions.
We use the recently introduced TERN architecture as an image-sentence features extractor. It is designed for producing fixed-size 1024-d vectors describing whole images and sentences, as well as variable-length sets of 1024-d vectors describing the various building components of the two modalities (image regions and sentence words respectively). All these vectors are enforced by the TERN design to lie into the same common space.
Our experiments show interesting preliminary results on the explored methods and suggest further experimentation in this important research direction.
\end{abstract}

\begin{IEEEkeywords}
deep features, information retrieval, cross-modal retrieval
\end{IEEEkeywords}

\section{Introduction}
In a cross-modal retrieval scenario, the elements in a database may pertain to arbitrary modalities (they could be images, videos, audio, or text); an interesting use-case arises when the query is a sentence expressed in natural language, while the retrieved elements are the images most related to the given query sentence. In this case, we are referring to the \textit{sentence-image} (or simply \textit{image-}) retrieval scenario. 
Conversely, in the \textit{image-sentence} (or simply \textit{sentence-}) retrieval an image is used as a query to find the most relevant sentences. 
These are interesting applications nowadays, where textual and visual search engines should be flexible, user-friendly, and easy to query.

Usually, the query objects and the elements to be retrieved are represented as vectors in a particular common space where a similarity function is defined. In cases where the query has the same modality as the retrieved elements, the same mapping function is sufficient to transform both the query and all the elements of the database into the same vector space where nearest-neighbor or range-query searches can be performed. When instead the query and the retrieved elements pertain to different modalities, it is not trivial in the general case to vectorize them so that in the end they lie into the same space.  

Recently, AI research gave a huge boost to the multi-modal processing of images and texts and proposed elegant solutions for the problem of bringing closer representations from the visual and textual modalities.
In fact, in the last years, the great advances in AI research brought to life interesting applications both in computer vision and in natural language processing worlds. In particular, deep neural networks demonstrated incredible performances in the processing of complex structured and non-structured data and they reached state-of-the-art results in many language and vision tasks. The joint analysis of images and texts gave birth to interesting applications, such as image captioning \cite{xu2015showattendtell,anderson2018butdimagecaptioning}, image-text matching \cite{li2019,vsepp2018faghri,lee2019,lu2019vilbert}, and weakly-supervised region-word alignment \cite{karpathy2015alignment,lee2018stackedcrossattention}.

Among all these tasks, image-text matching is the most interesting one in the context of image-sentence retrieval. Image-text matching neural architectures usually employ two different feature extraction pipelines able to produce vectors for the two modalities in a common representation space. Given a similarity function (e.g., cosine-similarity), they try to maximize the vector similarity between matching image-sentence pairs while minimizing it for non-matching pairs.

The matching performance is usually measured using retrieval metrics (Recall@K \cite{vsepp2018faghri,li2019,lu2019vilbert,lee2019} or NDCG \cite{carrara2018pictureit,messina2020transformer}), but the ranking is performed by sequentially scanning the entire dataset. Although this evaluation procedure is useful for quantifying the matching capability of the neural network, it is useless in real cross-modal retrieval scenarios, where images and sentences can scale up to millions of instances and the queries need to be solved in few milliseconds.
These issues motivate the research of methods able to produce multi-modal vectors that can be indexed, possibly using already existing structures such as inverted indexes, which exhibit high efficiency as shown in \cite{amato2014mifile}.

In this work, we perform an extensive evaluation of the procedures aimed at transforming non-indexable dense features in representations suitable for inverted indexes, to pave the way towards efficient yet effective cross-modal retrieval using deep features. For this reason, our experiments are directed towards a detailed evaluation of the effectiveness of the proposed features, although the long term path for this research is highly driven by efficiency concerns.

We use the TERN \cite{messina2020transformer} architecture as a visual-textual feature extractor. The TERN architecture can produce both global descriptors (a single fixed-sized vector) for images and sentences, together with fine-grained descriptions (a variable set of fixed-sized vectors) of image regions and words, all in the same common space, as shown in Figure \ref{fig:tern_architecture}. This provides us the opportunity to explore different aggregation and sparsification techniques: deep permutations \cite{amato2016deeppermutations} or scalar quantization \cite{amato2019large} when processing the global descriptors, and a variation of the Bag of Words model, that we call Bag of Concepts, when dealing with images and sentences as sets of concepts.


\section{Related Work}
\subsection*{Deep features for Information Retrieval}
With the advent of deep learning, many works introduced new classes of multimedia descriptors for information retrieval. In particular, focusing on the image retrieval world, deep features were found to be valid and cheaper alternatives to local features like SIFT, ORB, BRIEF, etc.

The basic idea behind image descriptors produced by a deep neural network is to use the neural activations of the last layers of an architecture trained on some specific task (e.g. image classification). The activations from the last layers carry a highly semantic representation of the image content that demonstrated very nice behaviors when used for image retrieval purposes \cite{tolias2016rmac,donahue2014decaf,razavian2014cnnfeatures}.

Recently, some works \cite{amato2016deeppermutations,amato2019large,jain2017subic,liu2018cnnindexing} tackled the problem of indexing this kind of features. They kept into consideration their high-dimensionality and their non-sparse nature, which poses major challenges when common indexing methods like inverted lists are employed.

In particular, \cite{amato2016deeppermutations} applied the permutation approach \cite{amato2014mifile} to deep features, where the vector elements are permuted based on the activation strength. 
Differently, \cite{amato2019large} tried to apply the deep permutation idea, together with a scalar quantization approach, to obtain a surrogate text representation that could be used in already existing document search engines such as Elasticsearch.

\subsection*{Bag of Words}
Our work goes into the direction of considering images and sentences as sets of basic elements we generically call \textit{concepts}.
From this perspective, our work lays its foundations on the Bag of Words (BoW) model, or Bag of Visual Words (BoVW) if we are referring to images \cite{sivic2003bow,Jegou2010bagoffeatures}. 

In literature, the BoVW model is used in close contact with local handcrafted features like SIFT, where every image is represented as a variable-length set of fixed-dimensional features. BoVW approach creates a codebook of visual words that enables all the images to be described as sets of codewords drawn from the built visual codebook.
This representation is immediately suitable for inverted indexes since it natively produces sparse representations that avoid accessing every posting list. In \cite{liu2018cnnindexing}, the authors tried to adapt the same Bag of Words ideas to index deep convolutional features, by clustering global CNN features to obtain a visual dictionary. The reuse of already existing inverted list approaches is a common path towards efficient image retrieval.
However, these recent approaches based on deep features do not natively deal with images as a set of objects, as deep features usually carry a global description of the image.

Furthermore, these works do not deal with multi-modal representations. Searching and retrieving multi-modal elements is indeed an important feature in modern search engines.

\subsection*{Image-Text matching}
This work takes inspiration from recent advances in Computer Vision and Natural Language Processing literature. In particular, deep learning demonstrated impressive results in the field of image-text matching \cite{vsepp2018faghri,li2019,lu2019vilbert,lee2019}. In particular, self-attention mechanisms like the transformer encoder \cite{devlin2019bert,vaswani2017transformer} were successful in encoding sequences of words and set of image regions by understanding their context. Context is very important for precise matching, as it describes interactions between \textit{objects} constituting a multimedia element (e.g. words inside a sentence or image regions inside a picture) which help to discriminate different scenarios. We claim that the context is an essential ingredient even in the cross-modal retrieval setups, as their performance is strongly linked to the quality of the matching.
These architectures learn contextualized multi-modal representations by minimizing a hinge-based triplet-loss function between features produced from both the visual and the textual pipeline. The similarity between features is often measured using dot-product or cosine-similarity.

\begin{figure}[t]
    \centering
    \includegraphics[page=2,width=\linewidth]{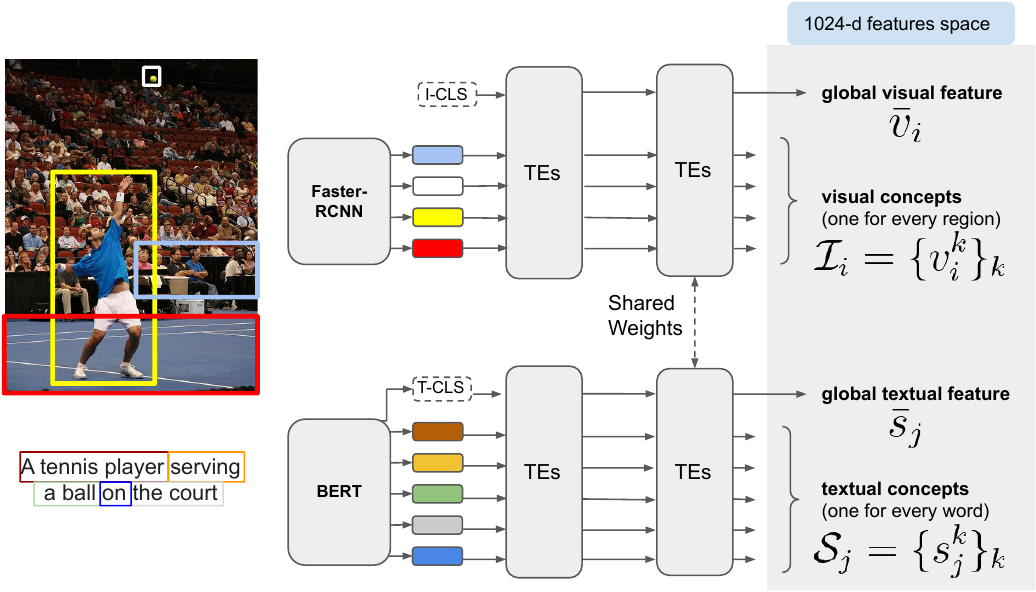}
  \caption{Overview of the TERN architecture. Please refer to \cite{messina2020transformer} for more details. The output consists of both global features summing up whole images and whole sentences, as well as contextualized concepts lying in the same common space.}
  \label{fig:tern_architecture} 
\end{figure}

This work is based on the achievements by \cite{messina2020transformer}. They developed an image-text matching architecture able to produce global and local descriptions of both images and sentences, all in the same common 1024-dimensional space (architecture summarized in Figure \ref{fig:tern_architecture}).

\section{Transforming TERN feature vectors}
Two routes are possible to reduce and sparsify the outputs of the TERN architecture (see Figure \ref{fig:tern_architecture}) for use with efficient data-structures such as inverted indexes. The first one deals with the sparsification of the $\bar{v}_i$ and $\bar{s}_j$ global vectors for images and sentences respectively, using approaches like deep permutations \cite{amato2016deeppermutations} or scalar quantization \cite{amato2019large}. The second possibility concerns the use of the fine-grained concepts extracted from the image regions $\mathcal{I}_i$ and the sentence words $\mathcal{S}_j$. In order to handle images and sentences described as sets of elementary components, we propose to construct a Bag of Concepts model which can be used to create a sparse description of images and sentences over a fixed-sized dictionary of reference concepts. Following, we will describe in detail these two methodologies.

\subsection{Dealing with Global Descriptions}
The 1024-d vectors $\bar{v}_i$ and $\bar{s}_j$ describing whole images and sentences can be transformed into a suitable vector indexable with inverted lists by applying the procedures described by \cite{amato2016deeppermutations,amato2019large}.
Inverted lists and surrogate text representation need and sparse and quantized representations, as they work with frequencies of appearance of terms in a dictionary. Both the scalar quantization and the deep permutation approaches try to obtain such representations from dense vectors of real numbers produced by neural networks.

Specifically, in the deep permutation approach every feature vector $\boldsymbol{v} \in \mathbb{R}^n$ is transformed by sorting the indexes
of the elements of $\boldsymbol{v}$ in descending order with respect to the corresponding element values. In this way, we can construct the permutation $\Pi_{{v}} = [\Pi_{\boldsymbol{v}}(1),...,\Pi_{\boldsymbol{v}}(n)]$ of the feature vector $\boldsymbol{v}$ with respect indexes $\{1,...,n\}$ such that:

\begin{equation}
\forall i = 1,...,n-1, \qquad \boldsymbol{v}(\Pi_{\boldsymbol{v}}(i)) \geq \boldsymbol{v}(\Pi_{\boldsymbol{v}}(i + 1))
\end{equation}

where $\boldsymbol{v}(j)$ is the $j$-th element of $\boldsymbol{v}$.

So, for example, if we have $\boldsymbol{v} = [0.2, 0.4, 0.1, 0.3, 0.6]$, the resulting permutation vector would be $\Pi_{\boldsymbol{v}} = [5, 2, 4, 1, 3]$.

On the other hand, the scalar quantization approach applies the following transformation to the original feature vector: $\boldsymbol{v} \rightarrow \lfloor s\boldsymbol{v}\rfloor$, where $s$ is a scale factor and $\lfloor \cdot \rfloor$ is the floor operation.

Note that these representations are as dense as the original feature vectors, while inverted indexes need sparse representations to be efficient. 
For this reason, we sparsify the vectors obtained with deep permutation and scalar quantization approaches by keeping only the first $z$ higher values, while forcing all the others to be zero, as in \cite{amato2019large}

In both cases, we measure the similarity between the transformed vectors using the standard cosine-similarity function.

\subsection{The \textit{Bag of Concepts} model}

The TERN architecture provides us a set of concepts describing images and sentences. In the case of images, every salient region carries a concept, while for sentences the concepts are associated with single words.
We are given a variable set of concepts $\mathcal{I}_i$  for every image $i$, and a variable sequence of concepts $\mathcal{S}_j$ for every sentence $j$.
The key idea is to produce a codebook of concepts so that both images and sentences can be described as a set of codewords drawn from a common dictionary.

\subsubsection{Creating the codebook}

The codebook can be produced by collecting a large amount of visual and textual concepts from the training set and then performing clustering as in the standard Bag of Visual Words model.
For this reason, we produce a large set of mixed visual and textual concepts:
\begin{align*}
    \mathcal{C} = \bigcup_i \mathcal{I}_i \cup \bigcup_j \mathcal{S}_j
\end{align*}
We downsample $\mathcal{C}$ so that $|\mathcal{C}| \simeq \text{100k concepts}$.

At this point, kmeans is used to produce $p$ clusters. The $p$ centroids represent our codebook of concepts. 
Given that the word and the visual word spaces correspond, it is also possible to create a common codebook by using only the textual words from all the sentences. If we follow this methodology, we can consider the top $p$ most common words appearing in all the $\mathcal{S}_j$ of the training set that are also present in the English dictionary and which are not stop-words.

\subsubsection{Inference}
Given a codebook, it is possible to produce the proper encoding for the images and sentences of the test set using the built dictionary. 
In the following discussion, we refer to the set of concepts $\mathcal{I}_i$ coming from images, although the same holds symmetrically for the set of concepts $\mathcal{S}_j$ from sentences.
If we use hard-assignment, we can proceed as follows: given the set of concepts $\{\boldsymbol{v}_i^k\}_{k\in\{1..n_i\}}$ from an image $i$, we can encode the image by finding, for every concept $\boldsymbol{v}_i^k$, the index of the nearest centroid (using L2 distance). Thus, in output, we obtain a set of $k$ codewords $\tilde{\mathcal{I}}_i = \{\tilde{v}_i^k\}_{k\in\{1..n_i\}}$, where every element $\tilde{v}_i^k$ of this set is no more a 1024-d vector but an integer representing the index of the nearest centroid.
The final representation for the image $i$ is obtained by computing the histogram $\boldsymbol{h}_i$ over the integer values contained in $\tilde{\mathcal{I}}_i$.
$\boldsymbol{h}_i$ has $p$ buckets and it is therefore a $p$-dimensional vector. It is already very sparse, with a maximum number $n_i$ of non-zero elements, where $n_i << p$.

\begin{figure}[t]
    \centering
    \includegraphics[page=1,width=\linewidth]{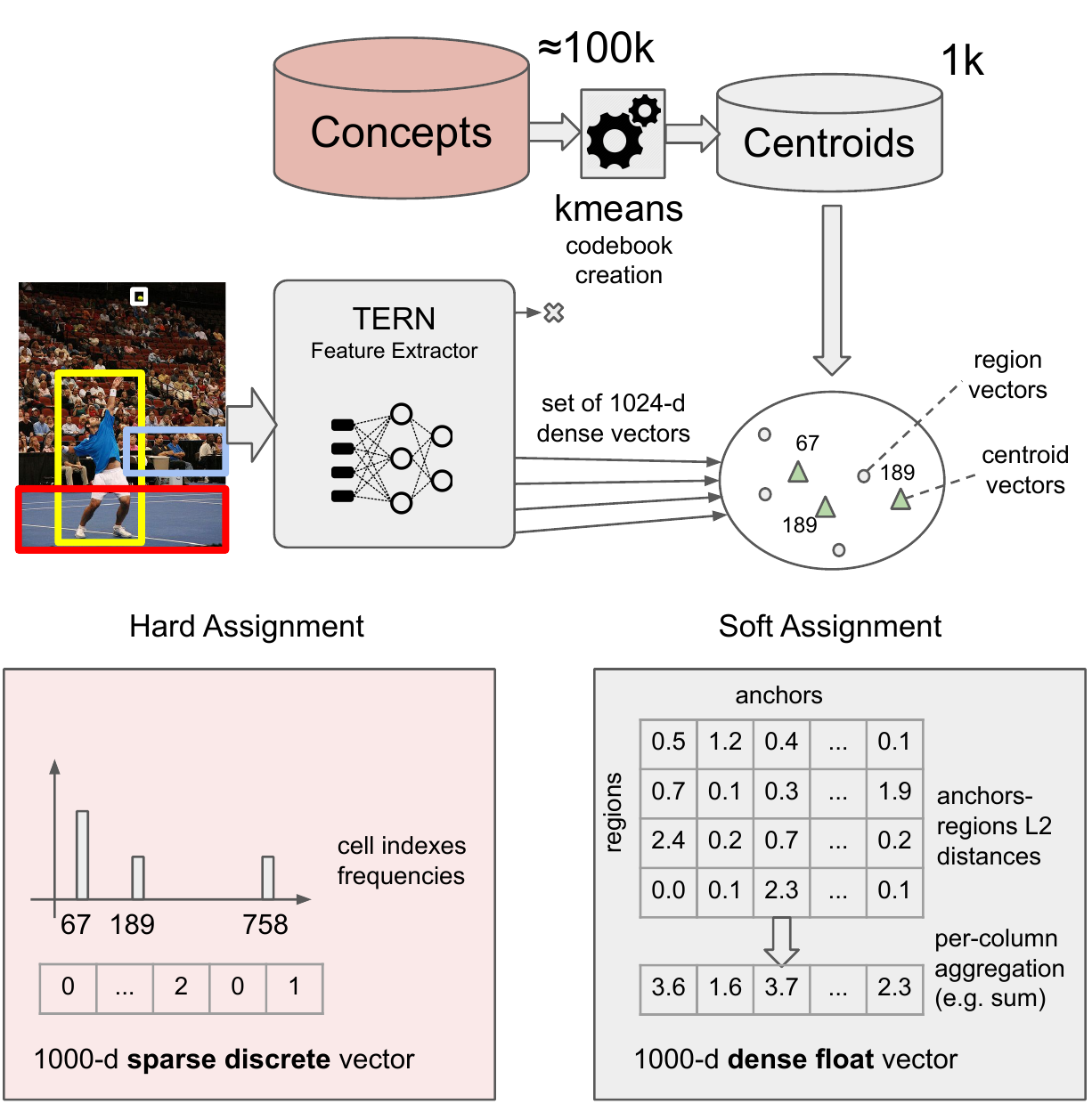}
  \caption{Overview of the Bag of Concept (BoC) model, in the case of image concepts inference. Both soft and hard assignment versions of BoC obtain fixed-sized descriptions from sets of possibly contextualized regions extracted from the image. Note that hard assignment already produces highly sparsified and discretized vectors.}
  \label{fig:boc_details} 
\end{figure}

The hard-assignment methodology performs heavy approximations on the original concept vectors, due to the discretization phase that transforms a 1024-d vector to a single discrete codeword.
The soft-assignment methodology tries to solve this problem: soft-assignment is performed by replacing the 1024-d vector for the $k$-th concept of the $i$-th image  $\boldsymbol{v}_i^k$ with a $p$-dimensional vector of distances computed against all the $p$ centroids. In this way, we preserve the information regarding all the distances between every concept and all the centroids. 
The result of this operation is a matrix $\boldsymbol{D}_i$ of L2 distances with shape $n_i \times p$. We convert L2 distances to similarities by applying the common transformation $\boldsymbol{S}_i = \frac{1}{1 - \boldsymbol{D}_i}$.
In order to produce a fixed-length $p$ vector $\boldsymbol{a}_i$ describing the image we can aggregate the columns of $\boldsymbol{S}_i$, thus constructing $\boldsymbol{a}_i$ as $a_i^k = \text{aggr}_l \: s_i^{l,k}$, where the function $aggr(\cdot)$ is a symmetric function.
An example of the overall inference procedure in the case of an image is reported in Figure \ref{fig:boc_details}.

In our setup we experiment with both $\max(\cdot)$ and $\text{sum}(\cdot)$ aggregation functions. By using $\text{sum}(\cdot)$ we are summing together the similarity contributions for each image concept with respect to a certain centroid. If all the image concepts are near a given centroid the sum will be very high, thus informing that probably that particular concept is highly present in the image. Instead, if we employ $\max(\cdot)$, we gather information only from the nearest image concept for each given centroid. In this case, the maximum concept similarity to a given centroid will be low only if all the concepts are far away from that specific centroid.

The problem with soft-assignment is that the vector $\boldsymbol{a}_i$ is still dense and hence very inefficient from the point of view of an inverted index.
We can think of sparsifying this vector by acting on the rows of $\boldsymbol{S}_i$ before computing the aggregation: for every row of $\boldsymbol{S}_i$, we keep only the $z$ higher values, by setting to zero all the others.

In the end, we use $\boldsymbol{h}_i$ and the sparsified $\boldsymbol{a}_i$ vectors as features from the hard and soft-assignment Bag of Concepts model, and they are compared using cosine-similarity.

\renewcommand{\arraystretch}{1.2}
\newcolumntype{C}{>{\centering\arraybackslash}X}
\newcolumntype{R}{D{,}{\pm}{1.6}}
\newcolumntype{L}{>{\raggedright\arraybackslash}p{3cm}}
\begin{table*}[htbp]
\caption{Recall@K metrics for our experiments on both MS-COCO and Flickr30K datasets}
\begin{center}
\begin{tabular}{Lcccccccccccc}
\toprule
& \multicolumn{6}{c}{\textbf{MS-COCO} (5K images, 25K sentences)} & \multicolumn{6}{c}{\textbf{Flickr30K} (10K images, 50K sentences)} \\
\cmidrule(lr){2-7} \cmidrule(lr){8-13}
& \multicolumn{3}{c}{Image Retrieval} & \multicolumn{3}{c}{Sentence Retrieval} & \multicolumn{3}{c}{Image Retrieval} & \multicolumn{3}{c}{Sentence Retrieval} \\
\cmidrule(lr){2-4} \cmidrule(lr){5-7} \cmidrule(lr){8-10} \cmidrule(lr){11-13}
\textbf{Model} & \multicolumn{1}{c}{K=1} & \multicolumn{1}{c}{K=5} & \multicolumn{1}{c}{K=10}
& \multicolumn{1}{c}{K=1} & \multicolumn{1}{c}{K=5} & \multicolumn{1}{c}{K=10} & \multicolumn{1}{c}{K=1} & \multicolumn{1}{c}{K=5} & \multicolumn{1}{c}{K=10} & \multicolumn{1}{c}{K=1} & \multicolumn{1}{c}{K=5} & \multicolumn{1}{c}{K=10} \\
\midrule
\multicolumn{13}{c}{\textit{Global Features}} \\
\midrule
TERN \cite{messina2020transformer} & 28.7 & 59.7 & 72.7 & 38.4 & 69.5 & 81.3 & 13.1 & 30.1 & 39.5 & 17.0 & 37.1 & 47.8\\
Deep Permutation & 28.7 & 59.8 & 72.7 & 38.5 & 69.6 & 81.3 & 13.2 & 30.1 & 39.5 & 17.1 & 37.1 & 47.9 \\
Scalar Quantization & 28.7 & 59.8 & 72.7 & 38.4 & 69.6 & 81.3 & 13.1 & 30.1 & 39.5 & 17.0 & 37.1 & 48.0 \\
\midrule
\multicolumn{13}{c}{\textit{Bag of Concepts - Hard Assignment}} \\
\midrule
No context & 3.5 & 14.0 & 24.2 & 4.0 & 14.7 & 23.1 & 0.8 & 2.9 & 4.9 & 1.1 & 3.9 & 6.1 \\
With context & 7.5 & 29.9 & 43.6 & 8.3 & 27.3 & 41.4 & 1.4 & 5.5 & 9.3 & 1.8 & 6.2 & 9.5 \\
No cont., no stop-words & 3.4 & 13.6 & 23.6 & 4.0 & 14.2 & 22.3 & 0.8 & 2.9 & 4.9 & 1.1 & 3.6 & 5.8 \\
With cont., no stop-words & 6.7 & 27.5 & 42.0 & 6.9 & 25.0 & 39.0 & 1.3 & 5.1 & 8.8 & 1.6 & 5.8 & 9.1 \\
No cont., no stop-words (inference) & 3.1 & 12.8 & 21.8 & 3.7 & 12.9 & 20.9 & 0.8 & 2.7 & 4.5 & 0.9 & 3.0 & 5.0 \\
English dict. & 2.7	& 9.4 & 16.4 & 2.8 & 10.0 & 16.5 & 1.1 & 4.2 & 7.1 & 1.3 & 4.7 & 7.7 \\
\midrule
\multicolumn{13}{c}{\textit{Bag of Concepts - Soft Assignment}} \\
\midrule
MAX-aggr No-cont. & 25.4 & 54.2 & 67.4 & 32.0 & 63.6 & 76.2 & 10.1 & 24.1 & 32.6 & 13.2 & 30.3 & 40.5\\
SUM-aggr No-cont. & 25.7 & 54.5 & 67.4 & 32.7 & 64.4 & 77.0 & 10.3 & 24.7 & 33.1 & 14.1 & 32.5 & 42.5 \\
MAX-aggr W-cont. & 26.8 & 57.0 & 70.4 & 35.1 & 65.1 & 77.9 & 10.6 & 25.5 & 34.1 & 14.1 & 32.4 & 42.3\\
SUM-aggr W-cont. & 27.2 & 57.4 & 70.4 & 34.9 & 65.9 & 78.4 & 10.7 & 25.8 & 34.4 & 14.6 & 33.1 & 43.5\\
\bottomrule
\end{tabular}
\label{tab:results}
\end{center}
\end{table*}

\section{Experiments}

In our experiments, we evaluated the retrieval effectiveness of the produced features. We experimented both with global features (transformed using deep permutation or scalar quantization) and local contextualized features, processed using the Bag of Concepts model. 
We measured the retrieval effectiveness using the MS-COCO and Flickr30K datasets, which come with 5 human-written captions describing each image.  
For MS-COCO, we used the 5k test images from the split introduced by \cite{karpathy2015alignment}, together with the associated 25k sentences. Regarding Flickr30K, we sampled 10k images and the corresponding 50k sentences from the training set, given that Flickr30K was not used for the training procedure of the TERN feature extractor. 

\begin{figure*}[t]
\begin{subfigure}[b]{0.3\textwidth}
\centering
\includegraphics[width=\linewidth]{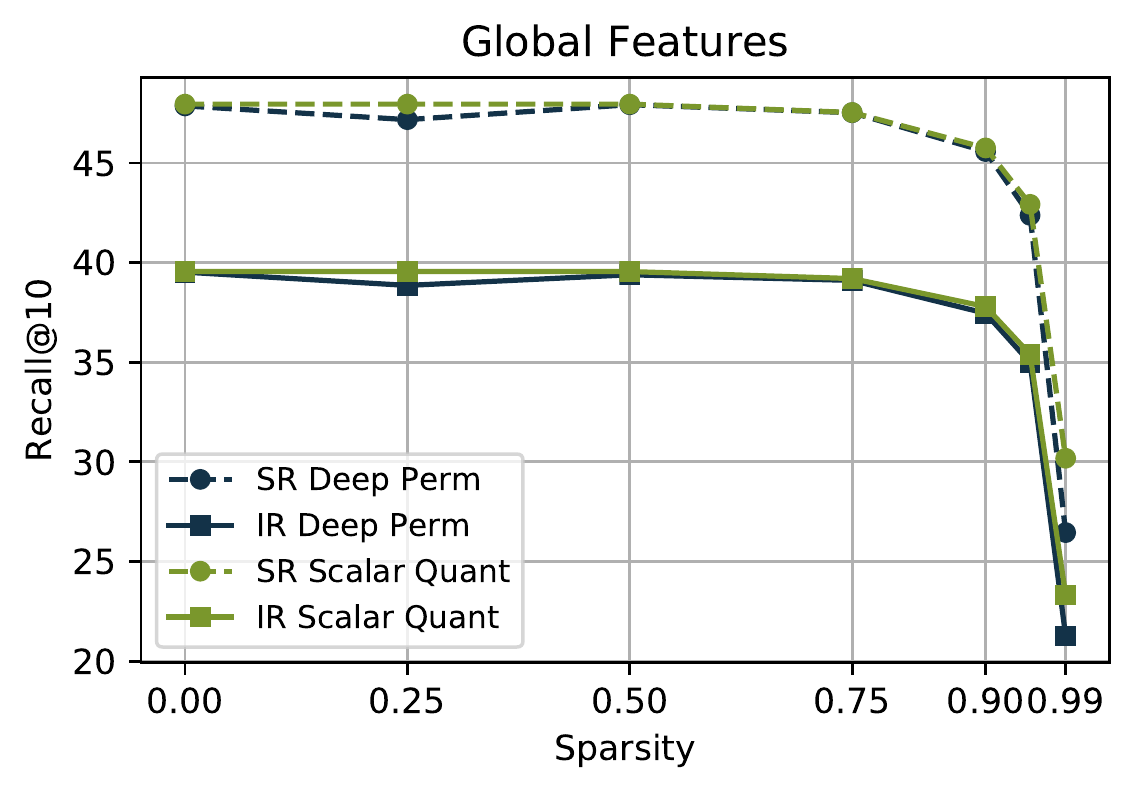}
\caption{Sparsity on Global Features}
\label{fig:sparsify_global}
\end{subfigure}
\begin{subfigure}[b]{0.3\textwidth}
\centering
\includegraphics[trim={0 0 0 0},clip,width=\linewidth]{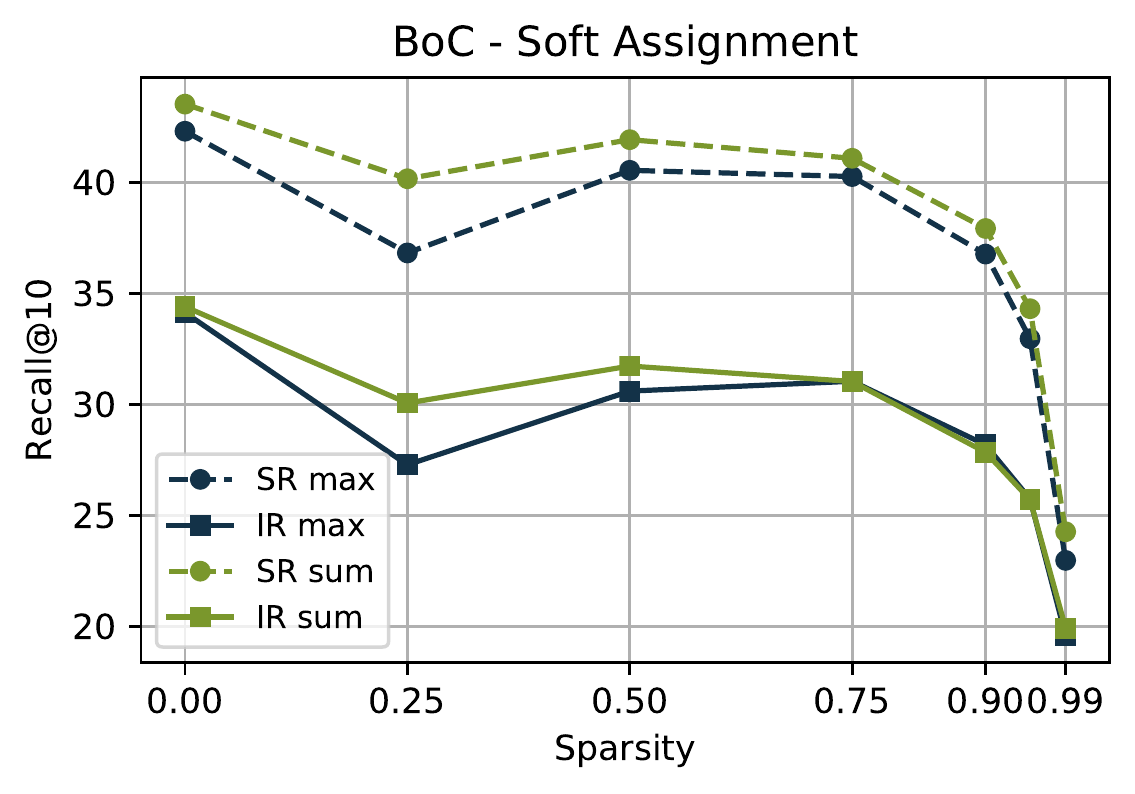}
\caption{Sparsity on BOW Soft Ass. Features}
\label{fig:sparsify_soft_assignment}
\end{subfigure}
\begin{subfigure}[b]{0.4\textwidth}
\centering
\includegraphics[width=\linewidth]{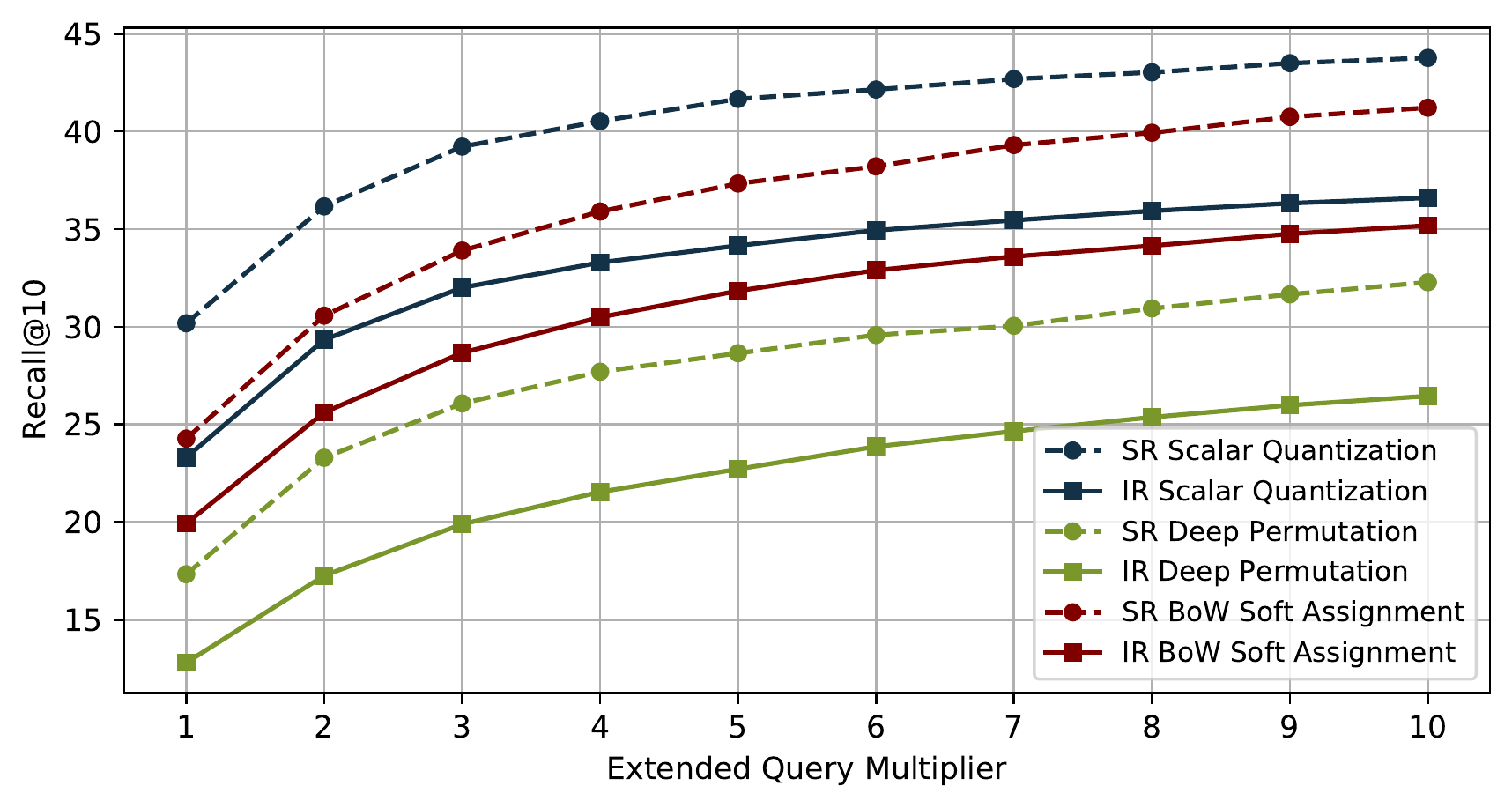}
\caption{Re-ranking effect with sparsity=0.99}
\label{fig:reranking} 
\end{subfigure}

\caption{(a) and (b): Trend of the Recall@10 metric while increasing the vector sparsity. Sparsity is reported as the fraction of vector components that are zeroed out ($\frac{z}{p}$ for BoC and $\frac{z}{2d}$ for global features, where $p=1000$ and $d=1024$ in our experiments). 
(c): Reranking results in terms of Recall@10 varying the extended query multiplier $R_m$. We search for the first $R_m \cdot 10$ items using the approximated method and then we rerank them using the original feature vectors. All the charts report the curves for both image retrieval (IR) and sentence retrieval (SR), and they are collected on the 10K images (50K sentences) from the Flickr30K dataset.}
\label{fig:sparsity}       
\end{figure*}

We used the Recall@K metric for evaluating the retrieval abilities of the collected features. The Recall@K is employed in many previous image-text retrieval works \cite{vsepp2018faghri,li2019,messina2020transformer}, and it measures the percentage of queries able to retrieve the correct item among the first $k$ results.

Concerning deep permutations and scalar quantization approaches, we followed \cite{amato2019large}: we first pre-processed the feature by applying a \textit{c-relu} operation, which concatenates the vector to its opposite, and then we clipped all negative values to zero. In this way, we obtained a vector containing only positive or zero integer elements. For the scalar quantization, we used a scale of 1,000.
Regarding the Bag of Concepts setup, we used a codebook of $p$=1,000 elements, either by clustering or by using the most frequent 1,000 words from the training set present also in the English dictionary. 

There are little variants of the Bag of Concepts that worth exploring.
Remember that the TERN architecture produces contextualized concepts that vary among different scenarios. If we want to avoid the contextualization effect during the clustering phase, we can forward regions and words one at a time inside TERN, so that it is impossible for the network to discern the different contexts. The de-contextualized scenario produces region and word features more similar in spirit to the hand-crafted local features like SIFT, which were highly de-contextualized. 
Furthermore, it is possible to exclude the stop-words during the clustering and/or indexing phases. This is often performed to avoid the noise produced by not-so-informative sentence words. The results of these experiments are reported in Table \ref{tab:results}. 



\subsection*{Results}

As it can be noticed from the first 3 rows from Table \ref{tab:results}, the application of deep permutations and scalar quantization does not change the essence of the ranking, as far as they are not sparsified. The Figures \ref{fig:sparsify_global} and \ref{fig:sparsify_soft_assignment} show how the features sparsification affects the overall performances on the 10K images from Flick30K. In particular, Figure \ref{fig:sparsify_global} shows that the results begin to diverge from the original TERN performance only when a massive sparsification is applied. In that case, the scalar quantization method produces features which are more resilient to a strong sparsification, both during image- and sentence-retrieval. 

Overall, the Bag of Concepts model can obtain very similar performances to the values reached by deep permutation and scalar quantization approaches, when soft-assignment is being used. 
Instead, the strong sparsification put in place by the hard assignment mechanism drastically lowers the overall effectiveness. Nevertheless, it can be noticed that we can always improve the overall Bag of Concepts performance by employing contextualized features. This is an important finding since it confirms that the context is a fundamental building block for understanding complex scenes: contextualized representations seem to have an important role even when considered as unordered sets without any structure, exactly like in the Bag of Concepts model.
Also, it is worth noting that the exclusion of stop-words both during the clustering and the indexing phases lowers the overall performances, indicating that they probably have a not negligible role in the pipeline.

When English words are used instead of the centroids from the kmeans clusters, we obtain performances worse than the non-contextualized kmeans case. This happens probably because these words natively lack context, as they are drawn from a dictionary and not computed from a representative training set. In fact, they could not be representative of the overall distribution of words in the MS-COCO dataset, although they are chosen among the 1,000 more frequent ones.

In Figure \ref{fig:reranking} we report the results after the re-ranking of the first $R_m\cdot K$ retrieved elements using the original global feature vectors in output directly from TERN, where $R_m$ is the extended query multiplier. For example, to build the Recall@5 metrics when $R_m=10$ we first retrieve the first $5 \times 10 = 50$ elements using the approximated method, and then we re-order them by computing the distances using the original feature vectors. The features used for the approximated search are sparsified with a sparsity factor of 0.99. 

The BoC with soft-assignment features cannot improve the results obtained by the deep permutation and scalar quantization ones, in case the sparsification is not applied (as shown in Table \ref{tab:results}). However, Figure \ref{fig:reranking} demonstrates that in case the sparsification factor is very high (0.99) the BoC soft-assignment features are more resilient and can almost bridge the performance gap with the scalar quantization features when the reranking multiplier is progressively increased. 

These results, in the end, show that the Bag of Concepts method has some interesting potential for efficient cross-modal retrieval, and further studies need to be performed in the direction of matching multi-media elements as complex sets of concepts.

\section{Conclusions and Future Work}
In this work, we employed the recently proposed TERN deep learning architecture as a multi-modal feature extractor. We leveraged the heterogeneous non-sparse output of this model to propose different solutions to the problem of generating indexable features for use in cross-modal retrieval systems.
Though not approaching the efficiency aspects, in this work we evaluated the retrieval effectiveness that these features exhibit if we apply common quantization - sparsification techniques to make them indexable in already existing efficient data-structures such as inverted indexes.

Experiments revealed that when the sparsification is not applied, the methodologies that process the global features still obtain overall better results than the Bag of Concepts ones.
When instead the sparsification is massive (e.g, when the sparsification factor reaches 0.99), the Bag of Concepts model with soft-assignment can obtain very competitive results during the re-ranking phase of the image retrieval scenario. This demonstrates the stability of the Bag of Concepts model in the presence of a critical sparsification.
Nevertheless, the Bag of Concepts, especially the sparsified soft-assignment methodology, deserves more attention as it is bounded to the concept features produced by the TERN architecture that, as of now, is not trained for producing exact region - words alignments.


In the near future, we seek to improve the Bag of Concepts model by including the production of effective concepts directly inside the TERN training loop. Also, the contextualization problem and the presence of inter-element connections between image regions and sentence words deserve more attention as much as indexing is concerned, since as of now the Bag of Concepts model cannot natively deal with relationships.

\section*{Acknowledgment}
This work was partially supported by “Intelligenza Artificiale per il Monitoraggio Visuale
dei Siti Culturali" (AI4CHSites) CNR4C program, CUP B15J19001040004 and by the AI4EU project, funded by the EC (H2020 - Contract n. 825619).



\bibliographystyle{IEEEtran}
\bibliography{biblio}

\end{document}